# A Heterogeneous Parallel Non-von Neumann Architecture System for Accurate and Efficient Machine Learning Molecular Dynamics

Zhuoying Zhao, Ziling Tan, Pinghui Mo, Xiaonan Wang, Dan Zhao, Xin Zhang, Ming Tao, and Jie Liu

*Abstract*—This paper proposes a special-purpose system to achieve high-accuracy and high-efficiency machine learning (ML) molecular dynamics (MD) calculations. The system consists of field programmable gate array (FPGA) and application specific integrated circuit (ASIC) working in heterogeneous parallelization. To be specific, a multiplication-less neural network (NN) is deployed on the non-von Neumann (NvN)-based ASIC (SilTerra 180 nm process) to evaluate atomic forces, which is the most computationally expensive part of MD. All other calculations of MD are done using FPGA (Xilinx XC7Z100). It is shown that, to achieve similar-level accuracy, the proposed NvN-based system based on low-end fabrication technologies (180 nm) is 1.6× faster and $10^2$-$10^3$× more energy efficiency than state-of-the-art vN-based MLMD using graphics processing units (GPUs) based on much more advanced technologies (12 nm), indicating superiority of the proposed NvN-based heterogeneous parallel architecture.

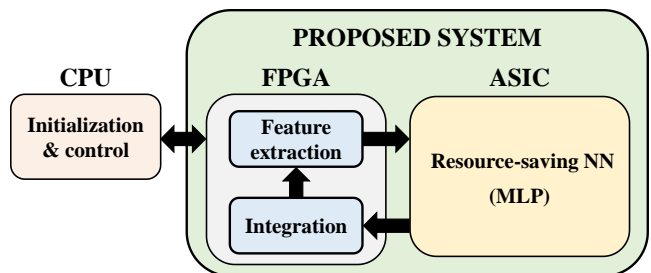

Fig. 1. Block diagram of the proposed MLMD computing system. The FPGA is responsible for feature extraction and integration, while the ASIC is in charge of implementing the resource-saving NN (a multilayer perceptron (MLP)). In addition, the proposed system requires a central processing unit (CPU) for initialization and control.

## I. INTRODUCTION

MOLECULAR dynamics (MD), as a computer simulation technique for complex systems modelled at the atomic level [1, 2], is widely used in many fields, including physics [3], chemistry [4], material science [5], semiconductor [6], nanotechnology [7], and biology [8], etc. The dilemma of accuracy versus efficiency has plagued the MD simulations for a long time. On one hand, density functional theory (DFT)-based ab-initio molecular dynamics (AIMD) is accurate, but its high computational cost limits its application in large systems [9-11]. Empirical force fields (EFF)-based classical MD (CMD) is efficient, but the manually-crafted EFF may deviate from the physical reality, leading to accuracy problems[12, 13].

The emerging machine learning (ML) MD (MLMD) has been proved to alleviate the long-standing dilemma between accuracy and efficiency [14-19]. By using the results of ab-initio calculations to train the neural network (NN) model, the MLMD can be several orders of magnitude faster than AIMD, while ensuring accurate MD calculations.

However, when running MD calculations, using von Neumann (vN) architecture computer is the only choice for most researchers, since vN architecture has been the dominating paradigm for many decades [20, 21]. Unfortunately, in the vN architecture, computing units (e.g., central processing unit (CPU) and graphics processing unit (GPU)) and storage units are separated from each other, so the majority (> 90%) of the total computation time and power consumption is actually consumed in the frequent data shuttling [22, 23]. Only a small fraction of time and power is used to perform useful arithmetic and logic operations. This is commonly known as the "vN bottleneck" (i.e., "memory wall bottleneck") [22, 23], seriously restricting the computing performance.

Although some special-purpose computers have been developed to accelerate MD calculations [24-26], they are all based on CMD, which makes their accuracy questionable in many important applications [27, 28]. Recently, by leveraging MLMD algorithms and NvN architecture, an MD computing system named NVNMD has been developed by Mo *et al* [29-31]. The NVNMD proves that the specially designed NvN-based MLMD computing system has higher computational speed and higher energy efficiency than the vN-based system by deploying on field programmable gate array (FPGA), providing a good hardware solution for accurate and efficient MLMD calculations. However, FPGA-based hardware

This work is supported by the National Natural Science Foundation of China (#61804049); the Fundamental Research Funds for the Central Universities of P.R. China; Huxiang High Level Talent Gathering Project (#2019RS1023); the Key Research and Development Project of Hunan Province, P.R. China (#2019GK2071); the Fund for Distinguished Young Scholars of Changsha (#kq1905012); the National Natural Science Foundation of China (#62104067); the National Natural Science Foundation of China (#62101182); the China Postdoctoral Science Foundation (#2020M682552). *(Corresponding author: Jie Liu; Ming Tao; Xin Zhang.)*

The authors are with the College of Electrical and Information Engineering, Hunan University, Changsha 410082, China (e-mail: jie_liu@hnu.edu.cn; tming@hnu.edu.cn; zhangxin2302@hnu.edu.cn).

architecture development limits the further improvement of the computational efficiency for MLMD computing systems due to its limited hardware resource and clock frequency, which come from FPGA being born in semi-custom application scenarios. By contrast, application specific integrated circuit (ASIC) has merits of more abundant hardware resources, higher clock frequency and lower power consumption than FPGA under the same process node [32, 33]. Thus, ASIC has better potential to further enhance the computational speed and energy efficiency of MLMD computing systems.

In this paper, a heterogeneous parallel ASIC-and-FPGA-based MLMD computing system (Fig. 1) is proposed and implemented, to boost the development of the NvN-based computing system for MLMD from the FPGA-based phase to the ASIC-based phase. The MLMD algorithm adopted in this paper consists of three modules, namely feature extraction, multilayer perceptron (MLP), and integration (see Section II), in which the MLP module is the most computationally expensive one. Our analysis shows that the execution of the MLP module accounts for the majority of the total calculation time by using either CPU or GPU machines, especially when the MLP size is large enough, the proportion can reach more than 90%. Therefore, deploying the MLP on an ASIC is the key to accelerate computing. Importantly, to reduce the physical resources for its hardware implementation, the MLP uses shift operations in place of multiplications during training stage, in conjunction with the specially designed lightweight activation function.

The resource-saving MLP is deployed on a carefully-optimized special-purpose NvN-based prototype, which is fabricated in SilTerra 180 nm process, with an area occupation of 1.73 mm$^2$ and a power dissipation of 1.9 W, to compute the atomic forces of a water molecule. Except for NN, the feature extraction module and integration module are implemented on a NvN-based Xilinx XC7Z100 FPGA. Overall, as shown in Fig. 1, the MLMD computing system consists of ASIC and FPGA, and an additional CPU is required for initialization and control. As a result, we demonstrate that the computational error of the proposed system is sufficiently small by measuring various physical properties of the water molecule. Moreover, compared to the state-of-the-art MLMD method relying on vN-based GPU with 12 nm process, the proposed system, measured at a low clock frequency of 25 MHz, returns 1.6× speedup and $10^2$-$10^3$× energy efficiency.

This work is a preliminary attempt to explore the substantial acceleration of NVNMD by ASIC, and the main purpose is to prove the feasibility. The remaining of the paper is organized as follows. Section II introduces MLMD. Section III discusses the optimization details of MLP module. Section IV introduces the architecture design and hardware implementation. Section V shows the measured results of the proposed system. Section VI and VII present a brief discussion and conclusion.

## II. MACHINE LEARNING MOLECULAR DYNAMICS

The procedure of MLMD calculations is introduced in this section. Firstly, the overall design of MLMD is briefly introduced (Section II-A). Then, the calculation flowchart is introduced in three modules (Section II-B): (i) the feature extraction module, (ii) the multilayer perceptron module, and (iii) the integration module.

### A. Overall Design

MLMD uses MLP to model energy or atomic force [15, 34, 35]. The input of the model is the local environment information related to the atomic position, also known as the feature. The output of the model is the energy or atomic force. If energy is the output, the force is calculated according to the energy derivative relative to the atomic position. In this work, MLP is used to predict the force directly, which can complete the MD calculations more efficiently.

MD is typically used to calculate the trajectories, $r_i(t)$ ($i$=1,2,…,$N_a$), of $N_a$ atoms in a system under certain conditions (e.g., temperature, pressure, etc.) [11]. Here, $t$ denotes time, $r_i$ denote atomic coordinates in the Cartesian space.

The schematic flow of MLMD adopted in this paper is presented in Fig. 2. Each MD step (time length $dt$) includes the following three modules. (i) Given the atomic coordinates at time $t$, $r_i(t)$, the features, $D_i(t)$, are computed. (ii) Using $D_i(t)$, the atomic forces, $F_i(t)$, are evaluated by MLP. (iii) Based on $F_i(t)$, through integrating Newton equation $F_i=m_i a_i$, $a_i=d^2 r_i/dt^2$, atomic coordinates at the next MD step, $r_i(t+dt)$, can be calculated. It should be noted that, to evaluate $F_i$, we only need to consider $N_{cut}$ neighbor atoms near atom $i$, whose locations, $r_j$, satisfy $|r_j - r_i| < r_{cut}$, where $r_{cut}$ is a cut-off radius, which can reduce the size of features. The full atomic trajectories can be obtained by repeating the above steps for a certain time.

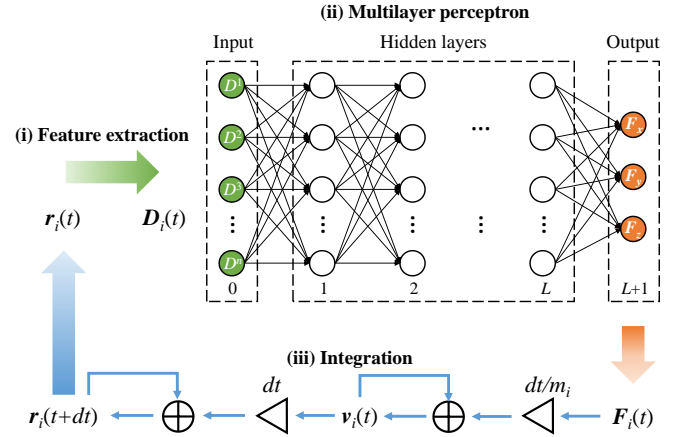

Fig. 2. Schematic flow adopted in this work to compute one MD step, which consists of three consecutive modules: (i) feature extraction, (ii) multilayer perceptron (MLP) force evaluation, and (iii) integration. Here, $t$ is time; $D_i(t)=(D^1, D^2, …, D^n)$ are features associated with atom $i$ ($i$=1,2,…, $N_a$); $n$ denotes the number of features; $N_a$ denotes the total number of atoms in the system; and $m_i$, $F_i$, and $v_i$ are mass, force, and velocity of atom $i$, respectively.

### B. Calculation Flowchart

*Feature Extraction Module*: As shown in the module (i) in Fig. 2, the atomic coordinates $r_i$ are converted into the features $D_i$, to preserve the translation, rotation and permutation symmetries [34].

*Multilayer Perceptron Module:* As shown in the module (ii) in Fig. 2, the MLP takes the features $D_i$ as the input to compute the atomic forces $F_i=(F_x, F_y, F_z)$. It is worth noting that the mapping from $D_i$ to $F_i$ is a complex high-dimensional problem, which is a challenge to compute both accurately and efficiently [14].

As we all know, MLP is mathematically capable of fitting arbitrarily complicated functions with arbitrary precision [36]. This mathematical conclusion provides a theoretical basis for us to map from $D_i$ to $F_i$ using MLP.

The MLP consists of $L+2$ layers, including an input layer (denoted as $l=0$), $L$ hidden layers (denoted as $l=1, 2, ..., L$), and an output layer (denoted as $l=L+1$). The output of the $l^{th}$ layer is

$$a_j^l = \phi\left(\sum_k w_{jk}^l a_k^{l-1} + b_j^l\right) \quad (1)$$

where $a_j^l$ is the output of the $j^{th}$ neuron of the $l^{th}$ layer; $w_{jk}^l$ is the weight connecting the $k^{th}$ neuron of the $(l-1)^{th}$ layer and the $j^{th}$ neuron of the $l^{th}$ layer; $b_j^l$ is the bias of the $j^{th}$ neuron of the $l^{th}$ layer; $\phi$ is the predefined nonlinear activation function; and $l=1, 2, …, L+1$. By setting the input of the MLP as features and the output as forces, the weights and biases can be trained using the DFT samples.

*Integration Module:* As shown in Fig. 2, the module (iii) computes atomic coordinates $r_i(t+dt)$ by using atomic forces $F_i(t)$ through

$$r_i(t+dt) = r_i(t) + v_i(t) \times dt \quad (2)$$

and

$$v_i(t) = v_i(t-dt) + \frac{F_i(t)}{m_i} \times dt \quad (3)$$

where $v_i$ and $m_i$ are the velocity and the mass of atom $i$, respectively [2].

## III. RESOURCE-SAVING NEURAL NETWORK

To achieve high computational efficiency with limited hardware resources, we adopt quantized NN (QNN) instead of continuous NN (CNN) (Section III-A). Two main optimizations are employed in QNN: Designing a lightweight nonlinear activation function (Section III-B); Using shift operation in place of multiplication operation to realize multiplication-less neural network (Section III-C). The effects of optimizations on the accuracy and hardware overhead are discussed in detail.

### A. Quantized Neural Network

Accurate calculations of the atomic forces by MLP is the key to perform accurate MD simulations. In traditional processors (e.g., CPUs and GPUs), MD simulations adopt high-precision floating-point numbers. However, continuous NN (CNN) based on floating-point numbers is very hardware resource-consuming in the implementation of the dedicated digital chip. Therefore, we use QNN to reduce the power and resource consumption of hardware design [37-39]. In the QNN, the weights and activations are quantized by using signed fixed-point numbers instead of floating-point numbers, so that integer arithmetic can be used to realize real number operations.

### B. Nonlinear Activation Function

The NN applied to regression problems usually uses hyperbolic tangent nonlinear activation function (i.e., $\tanh(x)$) [40], which is based on trigonometric function. If it is directly implemented, it will be very hardware resource-consuming [41]. Here, we design a hardware-friendly nonlinear activation function

$$\phi(x) = \begin{cases} 1 & x \geq 2 \\ x - \dfrac{x|x|}{4} & -2 < x < 2 \\ -1 & x \leq -2 \end{cases} \quad (4)$$

with fewer calculations. We can use the right shift to divide, for that the parameter in the denominator is the exponent of 2. The most complex operation in $\phi(x)$ is just multiplication. Compared with $\tanh(x)$, the hardware implementation of $\phi(x)$ is simpler, which can reduce the hardware overhead and improve the computational speed.

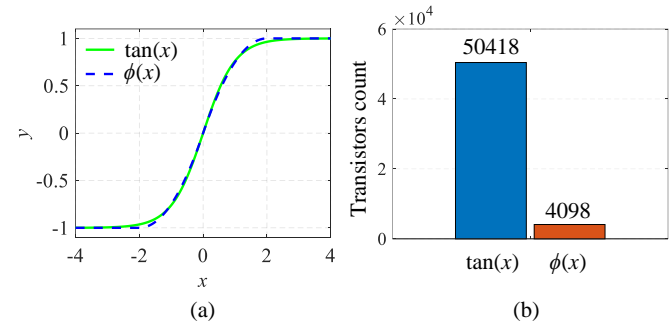

Fig. 3. (a) The curves of $\tanh(x)$ and $\phi(x)$. The $\tanh(x)$ and $\phi(x)$ are similar at the numerical value. (b) The number of transistors consumed by the $\tanh(x)$ and $\phi(x)$. The results are evaluated by using Synopsys Design Compiler (DC).

TABLE I
ACCURACY* COMPARISON OF NEURAL NETWORKS BASED ON TWO ACTIVATION FUNCTIONS

| Systems | $\tanh(x)$ | $\phi(x)$ | Difference** |
|---|---|---|---|
| Water | 25.04 | 24.83 | 0.21 |
| Ethanol | 29.33 | 29.84 | -0.51 |
| Toluene | 53.15 | 52.70 | 0.45 |
| Naphthalene | 46.45 | 46.63 | -0.18 |
| Aspirin | 74.85 | 75.20 | -0.35 |
| Silicon | 67.10 | 67.28 | -0.18 |

*The root mean square errors (RMSEs) of atomic forces (meV/Å) of six tested datasets. **Difference between $\tanh(x)$-based MLP and $\phi(x)$-based MLP.

*Accuracy:* The curves of $\tanh(x)$ and $\phi(x)$ are compared in Fig. 3(a). Obviously, $\tanh(x)$ and $\phi(x)$ are similar at the numerical value. In order to quantitatively analyze the influence of $\phi(x)$ on the fitting accuracy of MLP, using $\tanh(x)$ and $\phi(x)$, we train and test on six systems, including five molecule systems (i.e., water, ethanol, toluene, naphthalene, and aspirin), and one bulk system (i.e., silicon). The fitting accuracy of atomic forces obtained on the six datasets is shown in TABLE I. When compared against the DFT results, the atomic forces root mean square errors (RMSEs) (meV/Å) of $\tanh(x)$-based

MLP and that of $\phi(x)$-based MLP shows a very small difference, as indicated in the last column of TABLE I. In other words, replacing tanh($x$) with $\phi(x)$ will hardly bring accuracy loss.

*Hardware overhead:* We design and model the functions tanh($x$) and $\phi(x)$ by using the hardware description language (HDL) Verilog code [42], where tanh($x$) is implemented using the coordinate rotation digital computer (CORDIC) algorithm [43]. The register translation level (RTL) code is then converted into gate-level circuits using Synopsys logic synthesis tool Design Compiler (DC) [44], and the number of transistors consumed by the circuit is estimated based on the generated report. As indicated in Fig. 3(b), the tanh($x$) consumes 50418 transistors, while the $\phi(x)$ requires only 4098 transistors. It means that the hardware overhead of $\phi(x)$ is only 8% of that of tanh($x$), which greatly reduces the hardware overhead.

*C. Multiplication-Less Neural Network*

The key operation in the neural network (NN) is the multiply-accumulation (MAC). On traditional computing chips such as CPUs and GPUs, the calculation of MAC is very hardware resource-consuming and time-consuming. If the MAC operation is directly implemented in a dedicated digital circuit, it will also lead to large hardware overhead and power consumption [45-47].

In this paper, we propose a multiplication-less NN that reduces hardware overhead and power consumption by replacing multiplication operations with shift operations. Specifically, during training the model, we quantize the floating-point weights as sums of integer powers of 2 through

$$w_q = s(w) \cdot Q_K(w) \quad (5)$$

where, $w$ represents the floating-point weight; $w_q$ represents the quantized weight; $s(\cdot)$ is the sign function represented as

$$s(w) = \begin{cases} 1 & w > 0 \\ 0 & w = 0 \\ -1 & w < 0 \end{cases} \quad (6)$$

$Q_K(\cdot)$ is the quantization function given by

$$Q_K(w) = \begin{cases} Q_{K-1}\left(\max\left(|w| - Q(w), 0\right)\right) + Q(w) & K > 1 \\ Q(w) & K = 1 \end{cases} \quad (7)$$

where, $K$ stands for the number of integral powers of 2; max($x$, $y$) means take the larger between $x$ and $y$; $|\cdot|$ is the absolute value function; and the basis function,

$$Q(w) = 2^{\lceil \log_2(|w|/1.5) \rceil} \quad (8)$$

is used to quantize value to exponent of 2, where $\lceil \cdot \rceil$ means ceiling function to round a number to upper integer.

During inference stage, Eq. (5) can also be represented as

$$w_q = s \cdot \sum_{k=1}^{K} 2^{n_k} \quad (9)$$

where $s$ is the sign of weight $w$, obtained by Eq. (6); $n_k$ is the exponent, obtained by Eq. (7) and Eq. (8). Therefore, in the hardware implementation, the multiplication between weights and layer inputs will be replaced by a base-2 shift-sum operation, such that

$$w_q \cdot x_q = s \cdot \sum_{k=1}^{K} x_q \cdot 2^{n_k} = s \cdot \sum_{k=1}^{K} P(x_q, n_k) \quad (10)$$

where, $x_q$ represents the quantized layer input using fixed-point numbers; and

$$P(x, n) = \begin{cases} x \ll n & n > 0 \\ x \gg -n & n < 0 \\ x & n = 0 \end{cases} \quad (11)$$

is the shift function.

Obviously, after the above quantization design, the MAC operation in the NN will be completely replaced by the shift accumulate operation, which is very friendly to digital circuit implementation and can greatly reduce the complexity of the circuit.

Before quantitative analysis of accuracy and hardware overhead, some underlying conditions need to be stated. Firstly, the nonlinear activation function used in all models here is $\phi(x)$, based on the analysis in Section III-B. Secondly, we employ a pre-training strategy to improve the accuracy of QNN. Thirdly, for the same dataset, the corresponding CNN and QNN have the same size to ensure the fairness of comparison. Fourthly, for different datasets, the model size is different according to the complexity of the datasets. For the six datasets tested, the complexity of water, ethanol, toluene, naphthalene, aspirin, and silicon increases sequentially, resulting in a sequential increase in the corresponding network sizes.

*Accuracy:* Compared with the NN that solves the classification problem, the NN applied to the regression problem requires higher numerical precision and is more sensitive to the error caused by quantization. However, the accuracy and hardware cost of the NN are a dilemma with respect to the number of shifts (i.e., $K$ in Eq. (9)). On the one hand, ASIC implementation minimizes hardware cost with decrements of the number of shifts used to approximate a multiplication. On the other hand, NN tends to maximize accuracy with increments of the number of shifts. To find the appropriate $K$ value, we explore various evaluations for the following two models:

1) CNN: a baseline model using 32-bit floating-point numbers, which is a continuous standard MLP based on multiplication.
2) QNN: load the pre-trained CNN baseline model, quantify the weights according to Eq. (5)-(8), consider 5 different $K$ values (i.e., 1, 2, 3, 4 and 5), and train the model based on the pre-trained model.

The comparison between the accuracy of CNN and that of QNN is shown in Fig. 4. When $K$=1 or 2, the QNN has a serious accuracy loss, while from $K$=3, the loss tends to converge and the accuracy is gradually consistent with that of CNN. Furthermore, by calculating the ratio of RMSE of CNN to QNN, when $K$=3, the accuracy loss of QNN relative to CNN is between 6.5% and 12%.

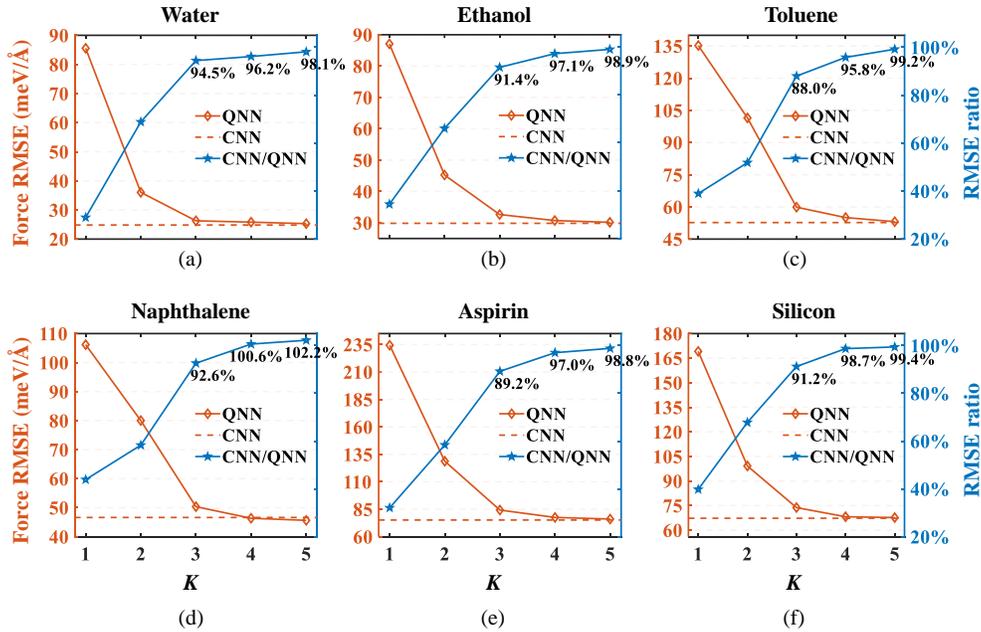

Fig. 4. Accuracy comparison between CNN and QNN. (a), (b), (c), (d), (e) and (f) are the results tested on the datasets of water, ethanol, toluene, naphthalene, aspirin and silicon, respectively. Each figure includes the force RMSE of CNN and QNN, as well as the RMSE ratio of CNN to QNN. With the increase of the number of shifts (i.e., $K$, whose value is 1 to 5), the accuracy of QNN gradually converges to that of CNN.

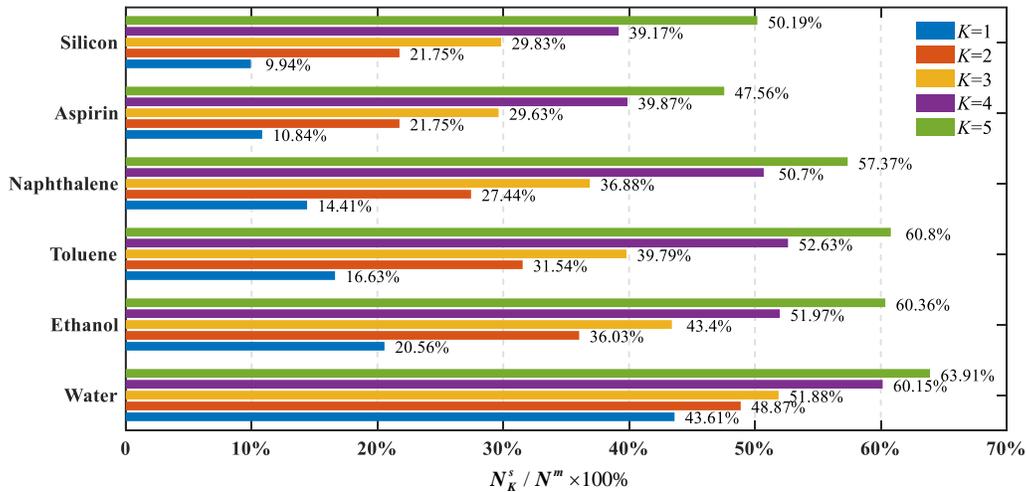

Fig. 5. The ratio of the number of transistors consumed by the SQNN to the FQNN. Here, $N^m$ denotes the number of transistors consumed by multiplication-based FQNN using 16-bit fixed-point quantization, and $N_K^s$ denotes the number of transistors consumed by shift-based SQNN. For each of the six datasets, the $K$ of SQNN is considered from 1 to 5.

*Hardware overhead:* Considering that the strategy widely used of deploying NN into hardware is to adopt fixed-point quantization schemes [48, 49], we perform 16-bit fixed-point quantization on CNN. We name the quantized CNN as FQNN to distinguish it from shift-based QNN (renamed as SQNN). For fair comparison, the layer input, bias and activation function of SQNN also use 16-bit fixed-point quantization except that the weight is quantized as the sum of powers of 2. By logically synthesizing the Verilog codes corresponding to SQNN and FQNN, the number of transistors consumed in the hardware implementation of SQNN and FQNN for six datasets is evaluated. We set the number of transistors consumed by multiplication-based FQNN with 16-bit fixed-point numbers as $N^m$ and the number of transistors consumed by shift-based SQNN as $N_K^s$. Then, the value, $N_K^s/N^m \times 100\%$, is calculated, as shown in Fig. 5. It can be obtained that the more complex the system is, the more hardware overhead can be saved by using SQNN. Combined with the analysis in terms of accuracy, for the value of $K$ when accuracy tends to converge (i.e., $K$=3), the SQNN can save about 50% to 70% of the hardware overhead relative to FQNN. At this time, increasing the $K$ (i.e., $K$=4 or 5) will not significantly improve the accuracy, but will increase the hardware cost by about 10% to 20%. Thus, $K$=3 is a more appropriate choice to the trade-off between accuracy and hardware overhead.

## IV. ARCHITECTURE DESIGN AND HARDWARE IMPLEMENTATION

This section introduces the architecture design and hardware implementation. The three modules of MLMD (Section II) are designed using non-von Neumann (NvN) architecture (Section IV-A). The MLP model applied to a single water molecule is designed and implemented in ASIC (Section IV-B). With the feature extraction module and the integration module implemented on FPGA, a heterogeneous parallel MLMD computing system is constructed by using ASIC and FPGA (Section IV-C).

*A. Non-von Neumann Architecture*

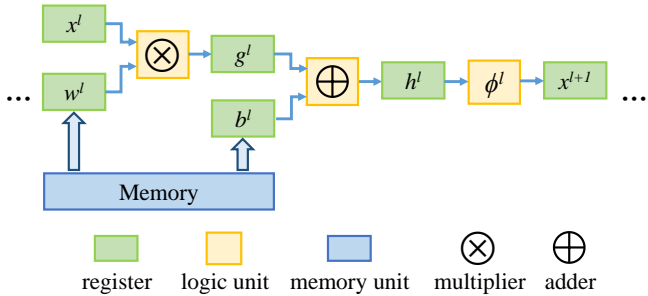

Fig. 6. Schematic of calculation step by adopting NvN architecture. Equation (1) is used as an example to shows the calculation process of NvN architecture. Here, $x^l$, $w^l$, $b^l$, and $\phi^l$ are input, weight, bias, and nonlinear activation function of the $l^{\text{th}}$ layer, respectively; $g^l = x^l \times w^l$; $h^l = b^l + g^l$.

As described in Section I, there are "memory wall bottlenecks" in vN architecture, which seriously restrict the improvement of MD computing performance. In this work, we adopt the NvN architecture, which is very important to improve the computational efficiency, especially to accelerate the MLP module with the highest computational density. The logical computing unit and the storage unit are integrated to avoid the repeated data shuttling. For instance, to calculate the $l^{\text{th}}$ layer of MLP, the weights $w^l$ and biases $b^l$ are stored in the locally distributed memory, directly participate in the calculation near it, and save the results in the nearby register. As shown in Fig. 6, the result $x^{l+1}$ of this layer is directly used as the input of the next layer, without saving the intermediate result to the off-chip memory. To compute a long MD trajectory of a particular material, $w^l$ and $b^l$ are only initialized once before MLP inference, and then kept unchanged during MD calculation of the full trajectories. We implement pipeline computing in the NvN architecture without data shuttling latency, so that the computational time is purely used for useful logic operations, improving the computational efficiency, and thus solving the "memory wall bottleneck" problem.

Similar to the MLP module, both feature extraction module and integration module are implemented using NvN architecture. All the calculations described in Section II-B are completed without back-and-forth data shuttling.

*B. Implementation of Multilayer Perceptron Chip*

We design and tape-out an ASIC for the MLP module, the most computationally intensive module among the three modules in the MLMD.

In order to verify the feasibility of our proposed method, we take the force prediction of a single water molecule as an example to train the model and implement it in ASIC. The process consists of 3 steps.

First, training samples are generated. AIMD is run to obtain the atomic trajectories, $r_{i(\text{DFT})}$, using density functional theory (DFT) code SIESTA [50]. The AIMD is run in a 2×2×2 nm supercell; Γ-point is used to sample Brillouin zone; double-zeta plus polarization (DZP) linear combination of atomic orbitals are used; plane wave cutoff is 100 Ry. To improve the MD calculation accuracy, a generalized gradient approximation is used to account for exchange-correlation effects [51]. The MD timestep $dt$ is set to as 2 fs. The atomic forces, $F_{i(\text{DFT})}$, at each MD step are calculated using the Hellman-Feynman theorem [52].

Second, an MLP model is trained. After the atomic coordinates $r_{i(\text{DFT})}$ are converted into features $D_i$ (see Section II-B), the MLP is trained using $D_i$ and $F_{i(\text{DFT})}$. Our training work is based on TensorFlow [53]. All simplified and quantized methods introduced in Section III are adopted in the training stage. Using 80% of the DFT samples as the training set and the remaining 20% as the test set, the atomic forces, $F_{i(\text{MLP})}$, predicted by the MLP can accurately reproduce the DFT results $F_{i(\text{DFT})}$. Here, we trained an MLP model to predict the forces on the hydrogen atom. The number of input neurons is 3, and the number of output neurons is 2. The model contains 2 hidden layers, and each hidden layer contains 3 neuron nodes. The forces on the oxygen atoms can be solved according to Newton's third law, to reduce the complexity of our design.

Third, the digital circuit of the MLP model is designed and implemented. The MLP mainly consists of two computationally expensive parts: weight matrix multiplication and evaluation of nonlinear activation function. In Section III-C, it has been introduced that the multiplication in MAC operation is replaced by shift operation, which is more suitable to realize processing in memory (PIM) in digital circuits, as the shift operation is cheaper in size and energy to be placed in or near the memory than multiplication. As shown in Fig. 7, we designed the matrix unit (MU) to realize the matrix multiplication between the weight and the layer input. In fact, the shift operation of $K$=3 is used. Therefore, the parameters we store are not the weight itself, but the corresponding shift parameters (i.e., $s$, $n_1$, $n_2$ and $n_3$ in Eq. (9)). The shift operation between each input and each weight is implemented using a shift unit (SU) (see Fig. 7), consisting of three shifters, an adder, and a symbol selector. After the shift accumulation between one row of the weight matrix and an input vector, it is added to bias. The result of addition passes through the activation function unit (AU). It can be seen from Fig. 7 that the AU only consists of two selectors, a multiplier, a shifter, and a subtracter, so the circuit implementation is simple. At the same time, the activation function circuit designed consumes less clock cycles than that of tanh($x$), because tanh($x$) needs more clock cycles to be iteratively solved. Each layer of MLP is implemented using the NvN architecture shown in Fig. 7. Using the SilTerra 180 nm process, the MLP chip is designed and tape-out. Fig. 8(c) shows the die micrograph of MLP chip, occupying 1.73 mm$^2$ die area. MLP chip on a printed circuit board (PCB) is exhibited in Fig. 8(b).

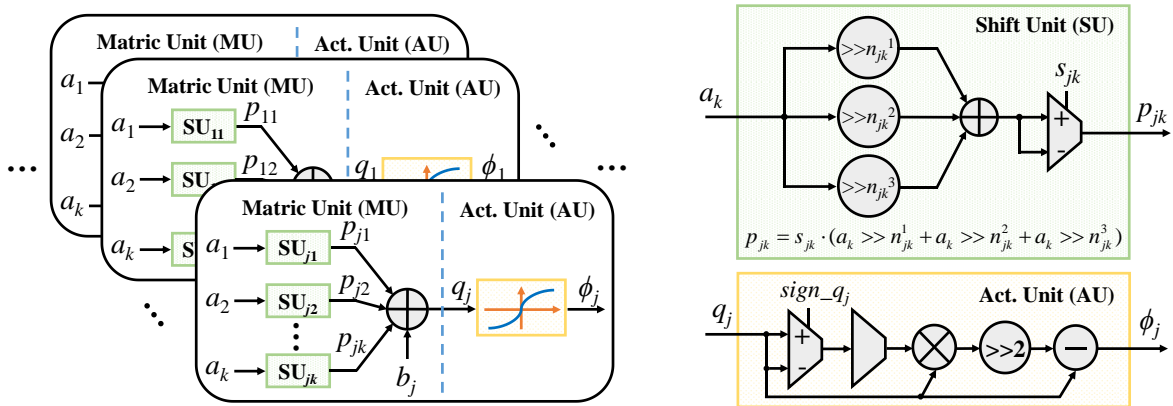

Fig. 7. Schematic implementation of the matrix multiplication and activation function of the $l^{th}$ layer in MLP, consisting of $j$ matric units (MU) and $j$ activation function units (AU). Each MU contains $k$ shift units (SU). Here, $j$ represents the number of neurons of the $l^{th}$ layer, and $k$ represents the number of neurons of the $(l\text{-}1)^{th}$ layer.

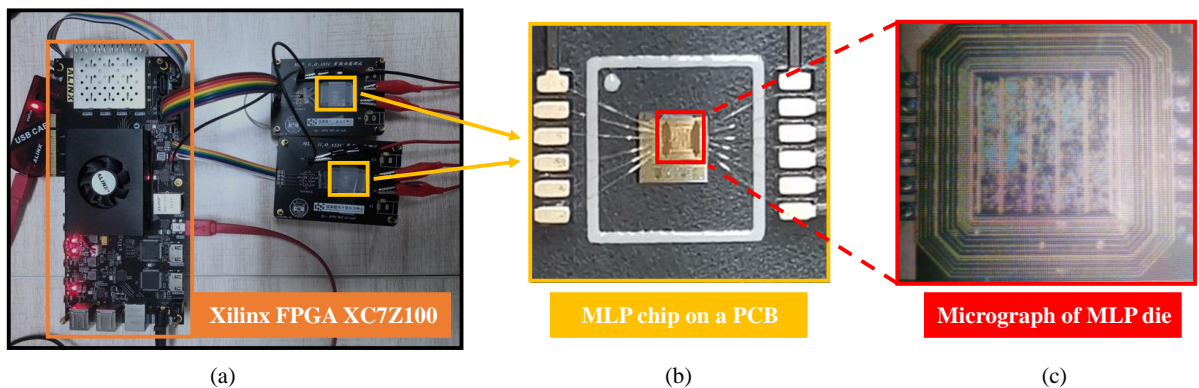

(a)     (b)     (c)

Fig. 8. (a) The heterogeneous parallel MLMD computing system. (b) MLP chip on a PCB. (c) Micrograph of MLP die.

*C. Heterogeneous Parallel System*

In this work, the MLP module is implemented based on the ASIC with NvN architecture, while the feature extraction module and integration module in MLMD calculations are implemented in the FPGA. The three modules of the MLMD all use signed 13-bit fixed-point numbers for operations, including 1 sign bit, 2 integer bits and 10 fractional bits. A heterogeneous parallel MLMD computing system for a single water molecule is shown in Fig. 8(a). The whole circuit consists of one Xilinx XC7Z100 FPGA and two MLP chips. The workflow of the system is: 1) The FPGA calculates the features of two hydrogen atoms in the water molecule; 2) The two sets of features are inputs to two MLP chips simultaneously, and the two chips work in parallel to predict the forces of two hydrogen atoms; 3) The two sets of forces are sent back to the FPGA, and the force of the oxygen atom is calculated based on Newton's third law. Using the forces, the integration process is performed to update the positional coordinates of the atoms. Repeating the process 1-3 to run multi-step MD calculations, the atomic trajectories can be obtained. Some physical properties can be further calculated from the trajectories, which will be introduced in detail in Section V.

## V. RESULTS

The measurement results of the proposed dedicated NvN-based MLMD computing system are presented in this section. First, it is verified that the MD calculation of the proposed system can achieve high-accuracy (Section V-A). Then, the computational speed (Section V-B) and energy efficiency (Section V-C) are quantitatively analyzed to demonstrate the high-efficiency.

*A. Accuracy*

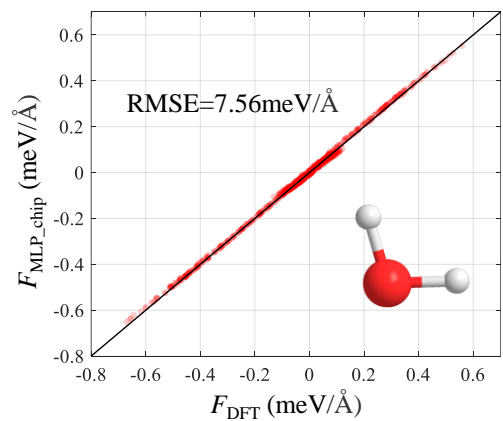

Fig. 9. The comparison between the forces of test set using the MLP chip and DFT for the water molecular. The RMSE is 7.56 meV/Å.

Reliable MD trajectories hinge on MLP's ability to evaluate the atomic forces accurately. Therefore, before the accuracy verification of MD calculation, we first test the function of the MLP chip. At the frequency of 25 MHz, as shown in Fig. 9, the atomic forces predicted by the proposed MLP chip is compared with the forces computed by the established DFT-based AIMD. The RMSE between the results of MLP chip and that of DFT is only 7.56 meV/Å.

The high-accuracy of the forces lays a solid foundation for reliable calculation of physical properties. Using the MD trajectories calculated by the proposed MLMD computing system, the structural properties (e.g., bond length and angle) and dynamic properties (e.g., vibration frequency) can be analyzed. As shown in TABLE II and Fig. 10, we measure the calculation results of the four methods, namely, the DFT results, vN-MLMD results, NvN-MLMD results and DeePMD [19] results. Among them, vN-MLMD and NvN-MLMD execute the same MLMD algorithm (see Section II). The difference is that vN-MLMD is deployed on the vN-based CPU (Intel Xeon E5-2696 v2), while NvN-MLMD uses MLMD computing system proposed in this work. As for DeePMD [19], it is an advanced and universal MLMD method. It is meaningful to compare our design with it. Furthermore, three relative errors are calculated, denoted as $Error^1$, $Error^2$ and $Error^3$, respectively. $Error^1$ shows that the vN-MLMD method achieves a very consistent effect with the DFT method, and the errors of all calculated properties are less than 1.18%, proving that MLMD has the similar high-accuracy to DFT method. $Error^2$ is more concerned, because it measures the accuracy of implemented NvN-MLMD. The results show that $Error^2$ does not exceed 1.06%, demonstrating that the proposed NvN-based work without sacrificing the high-accuracy of the MLMD. $Error^3$ shows the accuracy advantage of DeePMD, which is due to the fact that DeePMD uses a larger neural network and a more complex computing process compared with our work in terms of ensuring accuracy.

TABLE II
COMPARISON OF BOND LENGTH, ANGLE AND VIBRATION FREQUENCIES COMPUTED USING DIFFERENT METHODS

| Method | Bond length (Å) | H-O-H angle (°) | Vibration frequency (cm$^{-1}$) | | |
|---|---|---|---|---|---|
| | | | Symmetric stretching | Asymmetric stretching | Bending |
| DFT | 0.969 | 104.88 | 4007 | 4241 | 1603 |
| vN-MLMD | 0.968 | 104.90 | 4040 | 4291 | 1619 |
| NvN-MLMD | 0.968 | 104.85 | 4040 | 4274 | 1586 |
| DeePMD | 0.970 | 104.82 | 4003 | 4234 | 1599 |
| $Error^{1*}$ | 0.10% | 0.02% | 0.82% | 1.18% | 1.00% |
| $Error^{2*}$ | 0.10% | 0.03% | 0.82% | 0.78% | 1.06% |
| $Error^{3*}$ | 0.10% | 0.06% | 0.10% | 0.17% | 0.25% |

*Relative errors respectively computed by $Error^1 = \frac{|vN\text{-}MLMD - DFT|}{DFT} \times 100\%$, $Error^2 = \frac{|NvN\text{-}MLMD - DFT|}{DFT} \times 100\%$, and $Error^3 = \frac{|DeePMD - DFT|}{DFT} \times 100\%$.

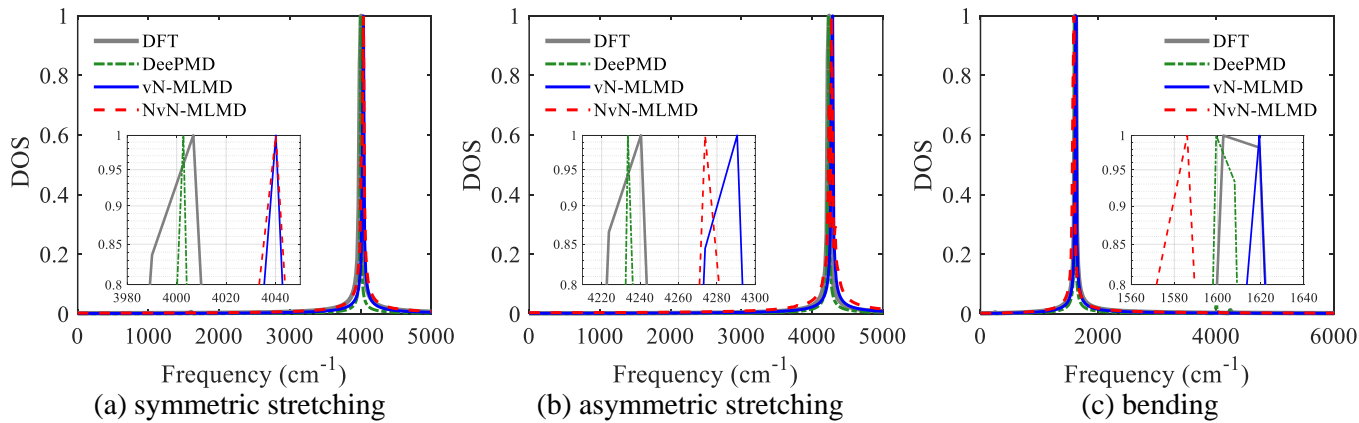

(a) symmetric stretching      (b) asymmetric stretching      (c) bending

Fig. 10. Vibration frequency of the water molecule. Here, DOS stands for the normalized density of states; the peak location indicates the vibration frequency. (a) symmetric stretching mode, (b) asymmetric stretching mode, and (c) bending mode of $H_2O$ vibrations are computed by using DFT, DeePMD, vN-MLMD and the proposed NvN-MLMD. The zoomed-out views of each plot are also shown.

TABLE III
COMPARISONS OF COMPUTATIONAL TIME COST AND ENERGY CONSUMPTION USING DIFFERENT METHODS

| Method | Hardware device | $S$ (s/step/atom) | $P$ (W) | $\eta = S \times P$ (J/step/atom) |
|---|---|---|---|---|
| DFT | CPU | 1.9 | 230 | $4.4 \times 10^2$ |
| vN-MLMD | CPU | $5.1 \times 10^{-4}$ | 45 | $2.3 \times 10^{-2}$ |
| DeePMD | CPU | $8.6 \times 10^{-5}$ | 152 | $1.3 \times 10^{-2}$ |
| DeePMD | CPU + GPU | $2.6 \times 10^{-6}$ | 250 | $6.5 \times 10^{-4}$ |
| NvN-MLMD | ASIC + FPGA | $1.6 \times 10^{-6}$ | 1.9 | $3.0 \times 10^{-6}$ |

## B. Speed

As shown in TABLE III, when applied to the MD calculation task of the water molecule, the computational speed of the proposed NvN-MLMD is about 6 orders of magnitude faster than the state-of-the-art DFT, and 1.6 times faster than the state-of-the-art GPU-based MLMD method, i.e., DeePMD [19]. It is worth noting that the results in Ref. [19] are obtained on an NVIDIA V100 GPU with 12 nm node, while the MLP chip in the proposed method adopts the 180 nm process. Due to the limitation of the MLP chip's process, the clock frequency adopted by the whole heterogeneous parallel system is 25 MHz. However, the clock frequency of the most of advanced commodity-level vN-based GPU/CPU can reach GHz-level [54, 55]. Although the processes used vary greatly and the clock frequency is about two orders of magnitude lower, the computational speed of the proposed NvN-MLMD is faster than that of the GPU/CPU-based MLMD method.

## C. Energy

The energy consumption $\eta$ is calculated by the formula $\eta=S\times P$, where $S$ represents the computational time cost and $P$ represents the power consumption. The measured total power consumption of the proposed NvN-MLMD is only 1.9 W, of which the power consumption of a single MLP chip is only 8.7 mW. As shown in TABLE III, the energy efficiency of the proposed system is $10^2$-$10^3\times$ higher than that of the state-of-the-art GPU-based MLMD method DeePMD [19].

## VI. DISCUSSION

The ASIC-based (180 nm process) method proposed in this paper is faster than the GPU (12 nm process), thanks to the adoption of the NvN architecture, which breaks the "memory wall bottleneck". It is foreseeable that NvN-MLMD will have faster computing speed when using more advanced process nodes.

The adoption of advanced process nodes has two main contributions to increasing computing speed. 1) The chips can reach clock frequencies of several GHz [54, 55], which means that, through purely boosting the clock frequency from 25MHz to several GHz, the computational speed can be directly accelerated by about 2 orders of magnitude (i.e., $A_1\approx 10^2$). 2) Higher intra-ASIC parallelization can be achieved in the same area due to higher integration of transistors in advanced processes. Take the 14 nm node as an example, it can be learned from Ref. [56] and Ref. [57] that the transistor integration of the 14 nm node is about 2 orders of magnitude higher than that of the 180 nm node. Therefore, it's anticipated that the computational speed could be enhanced by about 2 orders of magnitude (i.e., $A_2\approx 10^2$), by purely increasing the intra-ASIC parallelization. To sum up, the computational speed of the estimated NvN-MLMD would be around 4 orders of magnitude (i.e., $A_1\times A_2\approx 10^4$) faster than that of the proposed method in this paper. In other words, the computational time cost of the MLMD computing system could be reduced from $10^{-6}$ s/step/atom to around $10^{-10}$ s/step/atom, which shows great prospects of the NvN-MLMD.

For different MD tasks, if different NN models are used, the current ASIC design needs to be modified. Therefore, developing a universal architecture is an important work we are doing. For example, at the software algorithm level, we will deploy the MLMD algorithm that is widely applicable to different MD tasks. At the hardware architecture level, we will provide a variable NN size to meet the different needs of different tasks on the NN size.

## VII. CONCLUSION

In this work, a resource-saving and NvN-based MLP chip has been designed and implemented using SilTerra 180 nm process, to predict atomic forces. A heterogeneous parallel MLMD computing system has been proposed based on ASIC and FPGA. It is shown that, without compromising the high calculation accuracy, the proposed NvN-based MLMD achieves $1.6\times$ speedup and $10^2$-$10^3\times$ energy efficiency compared to the state-of-the-art vN-based MLMD method based on much more advanced process (12 nm). This paves the way for the development of next-generation NvN-based MLMD based on high-end fabrication technologies.

REFERENCES

[1] "Molecular dynamics." https://www.nature.com/subjects/molecular-dynamics (accessed.
[2] D. Frenkel and B. Smit, *Understanding molecular simulation: from algorithms to applications*. Elsevier, 2001.
[3] V. Bapst *et al.*, "Unveiling the predictive power of static structure in glassy systems," *Nature Physics,* vol. 16, no. 4, pp. 448-454, 2020.
[4] P. Bajaj, J. O. Richardson, and F. Paesani, "Ion-mediated hydrogen-bond rearrangement through tunnelling in the iodide–dihydrate complex," *Nature chemistry,* vol. 11, no. 4, pp. 367-374, 2019.
[5] F. Rao *et al.*, "Reducing the stochasticity of crystal nucleation to enable subnanosecond memory writing," *Science,* vol. 358, no. 6369, pp. 1423-1427, 2017.
[6] M. Shi, P. Mo, and J. Liu, "Deep neural network for accurate and efficient atomistic modeling of phase change memory," *IEEE Electron Device Letters,* vol. 41, no. 3, pp. 365-368, 2020.
[7] J. Liu *et al.*, "A sensitive and specific nanosensor for monitoring extracellular potassium levels in the brain," *Nature Nanotechnology,* vol. 15, no. 4, pp. 321-330, 2020.
[8] M. Karplus and G. A. Petsko, "Molecular dynamics simulations in biology," *Nature,* vol. 347, no. 6294, pp. 631-639, 1990.
[9] W. Kohn and L. J. Sham, "Self-consistent equations including exchange and correlation effects," *Physical review,* vol. 140, no. 4A, p. A1133, 1965.
[10] R. Car and M. Parrinello, "Unified approach for molecular dynamics and density-functional theory," *Physical review letters,* vol. 55, no. 22, p. 2471, 1985.
[11] D. Marx and J. Hutter, *Ab initio molecular dynamics: basic theory and advanced methods*. Cambridge University Press, 2009.
[12] W. L. Jorgensen, D. S. Maxwell, and J. Tirado-Rives, "Development and testing of the OPLS all-atom force field on conformational energetics and properties of organic liquids," *Journal of the American Chemical Society,* vol. 118, no. 45, pp. 11225-11236, 1996.
[13] J. Wang, R. M. Wolf, J. W. Caldwell, P. A. Kollman, and D. A. Case, "Development and testing of a general amber force field," *Journal of computational chemistry,* vol. 25, no. 9, pp. 1157-1174, 2004.
[14] J. Behler and M. Parrinello, "Generalized neural-network representation of high-dimensional potential-energy surfaces," *Physical review letters,* vol. 98, no. 14, p. 146401, 2007.
[15] N. Kuritz, G. Gordon, and A. Natan, "Size and temperature transferability of direct and local deep neural networks for atomic forces," *Physical Review B,* vol. 98, no. 9, p. 094109, 2018.
[16] H. Wang, L. Zhang, J. Han, and E. Weinan, "DeePMD-kit: A deep learning package for many-body potential energy representation and

molecular dynamics," *Computer Physics Communications,* vol. 228, pp. 178-184, 2018.

[17] L. Zhang, J. Han, H. Wang, W. Saidi, and R. Car, "End-to-end symmetry preserving inter-atomic potential energy model for finite and extended systems," *Advances in Neural Information Processing Systems,* vol. 31, 2018.

[18] W. Jia *et al.*, "Pushing the limit of molecular dynamics with ab initio accuracy to 100 million atoms with machine learning," in *SC20: International conference for high performance computing, networking, storage and analysis*, 2020: IEEE, pp. 1-14.

[19] D. Lu *et al.*, "DP train, then DP compress: model compression in deep potential molecular dynamics," *arXiv preprint arXiv:2107.02103,* 2021.

[20] J. Von Neumann, "First Draft of a Report on the EDVAC," *IEEE Annals of the History of Computing,* vol. 15, no. 4, pp. 27-75, 1993.

[21] "Electronic Numerical Integrator and Computer (ENIAC)." https://en.wikipedia.org/wiki/ENIAC (accessed.

[22] W. A. Wulf and S. A. McKee, "Hitting the memory wall: Implications of the obvious," *ACM SIGARCH computer architecture news,* vol. 23, no. 1, pp. 20-24, 1995.

[23] M. Horowitz, "1.1 computing's energy problem (and what we can do about it)," in *2014 IEEE International Solid-State Circuits Conference Digest of Technical Papers (ISSCC)*, 2014: IEEE, pp. 10-14.

[24] D. E. Shaw *et al.*, "Anton, a special-purpose machine for molecular dynamics simulation," *Communications of the ACM,* vol. 51, no. 7, pp. 91-97, 2008.

[25] D. E. Shaw *et al.*, "Anton 2: raising the bar for performance and programmability in a special-purpose molecular dynamics supercomputer," in *SC'14: Proceedings of the International Conference for High Performance Computing, Networking, Storage and Analysis*, 2014: IEEE, pp. 41-53.

[26] D. E. Shaw *et al.*, "Anton 3: twenty microseconds of molecular dynamics simulation before lunch," in *Proceedings of the International Conference for High Performance Computing, Networking, Storage and Analysis*, 2021, pp. 1-11.

[27] V. L. Deringer and G. Csányi, "Machine learning based interatomic potential for amorphous carbon," *Physical Review B,* vol. 95, no. 9, p. 094203, 2017.

[28] J. Zeng, L. Cao, M. Xu, T. Zhu, and J. Z. Zhang, "Complex reaction processes in combustion unraveled by neural network-based molecular dynamics simulation," *Nature communications,* vol. 11, no. 1, pp. 1-9, 2020.

[29] P. Mo *et al.*, "Accurate and efficient molecular dynamics based on machine learning and non von Neumann architecture," *npj Computational Materials,* vol. 8, no. 1, pp. 1-15, 2022.

[30] J. Liu and P. Mo. "The server website of NVNMD." http://nvnmd.picp.vip (accessed.

[31] J. Liu and P. Mo. "The training and testing code for NVNMD." https://github.com/LiuGroupHNU/nvnmd (accessed.

[32] I. Kuon and J. Rose, "Measuring the gap between FPGAs and ASICs," *IEEE Transactions on computer-aided design of integrated circuits and systems,* vol. 26, no. 2, pp. 203-215, 2007.

[33] A. De Vita, A. Russo, D. Pau, L. Di Benedetto, A. Rubino, and G. D. Licciardo, "A partially binarized hybrid neural network system for low-power and resource constrained human activity recognition," *IEEE Transactions on Circuits and Systems I: Regular Papers,* vol. 67, no. 11, pp. 3893-3904, 2020.

[34] L. Zhang, J. Han, H. Wang, R. Car, and E. Weinan, "Deep potential molecular dynamics: a scalable model with the accuracy of quantum mechanics," *Physical review letters,* vol. 120, no. 14, p. 143001, 2018.

[35] P. Mo, M. Shi, W. Yao, and J. Liu, "Transfer Learning of Potential Energy Surfaces for Efficient Atomistic Modeling of Doping and Alloy," *IEEE Electron Device Letters,* vol. 41, no. 4, pp. 633-636, 2020.

[36] K. Hornik, M. Stinchcombe, and H. White, "Multilayer feedforward networks are universal approximators," *Neural networks,* vol. 2, no. 5, pp. 359-366, 1989.

[37] G. Purushothaman and N. B. Karayiannis, "Quantum neural networks (QNNs): inherently fuzzy feedforward neural networks," *IEEE Transactions on neural networks,* vol. 8, no. 3, pp. 679-693, 1997.

[38] S. Gupta, A. Agrawal, K. Gopalakrishnan, and P. Narayanan, "Deep learning with limited numerical precision," in *International conference on machine learning*, 2015: PMLR, pp. 1737-1746.

[39] P. Gysel, J. Pimentel, M. Motamedi, and S. Ghiasi, "Ristretto: A framework for empirical study of resource-efficient inference in convolutional neural networks," *IEEE transactions on neural networks and learning systems,* vol. 29, no. 11, pp. 5784-5789, 2018.

[40] R. C. Minnett, A. T. Smith, W. C. Lennon Jr, and R. Hecht-Nielsen, "Neural network tomography: Network replication from output surface geometry," *Neural Networks,* vol. 24, no. 5, pp. 484-492, 2011.

[41] S. Marra, M. A. Iachino, and F. C. Morabito, "High speed, programmable implementation of a tanh-like activation function and its derivative for digital neural networks," in *2007 International Joint Conference on Neural Networks*, 2007: IEEE, pp. 506-511.

[42] S. Palnitkar, *Verilog HDL: a guide to digital design and synthesis*. Prentice Hall Professional, 2003.

[43] M. Garrido, P. Källström, M. Kumm, and O. Gustafsson, "CORDIC II: a new improved CORDIC algorithm," *IEEE Transactions on Circuits and Systems II: Express Briefs,* vol. 63, no. 2, pp. 186-190, 2015.

[44] "Synopsys Design Compiler." https://solvnet.synopsys.com/DocsOnWeb (accessed.

[45] M. Courbariaux, I. Hubara, D. Soudry, R. El-Yaniv, and Y. Bengio, "Binarized neural networks: Training deep neural networks with weights and activations constrained to+ 1 or-1," *arXiv preprint arXiv:1602.02830,* 2016.

[46] B. Wu *et al.*, "Shift: A zero flop, zero parameter alternative to spatial convolutions," in *Proceedings of the IEEE conference on computer vision and pattern recognition*, 2018, pp. 9127-9135.

[47] H. Chen *et al.*, "AdderNet: Do we really need multiplications in deep learning?," in *Proceedings of the IEEE/CVF conference on computer vision and pattern recognition*, 2020, pp. 1468-1477.

[48] D. D. Lin, S. S. Talathi, and V. S. Annapureddy, "Fixed Point Quantization of Deep Convolutional Networks," in *33rd International Conference on Machine Learning*, New York, NY, Jun 20-22 2016, vol. 48, in Proceedings of Machine Learning Research, 2016.

[49] J. Choi, Z. Wang, S. Venkataramani, P. I.-J. Chuang, V. Srinivasan, and K. Gopalakrishnan, "Pact: Parameterized clipping activation for quantized neural networks," *arXiv preprint arXiv:1805.06085,* 2018.

[50] J. M. Soler *et al.*, "The SIESTA method for ab initio order-N materials simulation," *Journal of Physics: Condensed Matter,* vol. 14, no. 11, p. 2745, 2002.

[51] J. P. Perdew, K. Burke, and M. Ernzerhof, "Generalized gradient approximation made simple," *Physical review letters,* vol. 77, no. 18, p. 3865, 1996.

[52] R. Gaudoin and J. Pitarke, "Hellman-Feynman operator sampling in diffusion Monte Carlo calculations," *Physical review letters,* vol. 99, no. 12, p. 126406, 2007.

[53] M. Abadi, P. Barham, J. Chen, Z. Chen, and X. Zhang, "TensorFlow: A system for large-scale machine learning," in *USENIX Association*, 2016.

[54] N. Corporation. "Nvidia Tesla V100 GPU Volta Architecture." White Paper 53. https://images.nvidia.cn/content/volta-architecture/pdf/volta-architecture-whitepaper.pdf (accessed.

[55] "Intel Core i9-10900K Processor." https://www.intel.com/content/www/us/en/products/sku/199332/intel-core-i910900k-processor-20m-cache-up-to-5-30-ghz/specifications.html (accessed.

[56] "International Technology Roadmap for Semiconductors 2.0 2015 edition." http://www.itrs2.net/itrs-reports.html (accessed.

[57] W. M. Holt, "1.1 Moore's law: A path going forward," in *2016 IEEE International Solid-State Circuits Conference (ISSCC)*, 2016: IEEE, pp. 8-13.